\pdfoutput=1

\documentclass[11pt]{article}

\usepackage[final]{acl}

\usepackage{times}
\usepackage{latexsym}

\usepackage[T1]{fontenc}

\usepackage[utf8]{inputenc}
\usepackage{enumitem}
\usepackage{microtype}

\usepackage{inconsolata}

\usepackage{graphicx}

\usepackage{amsmath,amssymb,amsfonts}
\usepackage{algorithmic}
\usepackage{graphicx}
\usepackage{textcomp}
\usepackage{xcolor}
\usepackage{todonotes}
\usepackage{booktabs}
\usepackage{url}
\usepackage{microtype}
\usepackage{multirow}
\usepackage{tabularx}
\usepackage{makecell}

\title{Vietnamese Automatic Speech Recognition: A Revisit}

\author{Thi Vu, Linh The Nguyen, Dat Quoc Nguyen\\
Qualcomm AI Research\thanks{Qualcomm Vietnam Company Limited. Qualcomm AI Research is an initiative of Qualcomm Technologies, Inc.}\\
\texttt{\{thivu, linhnt, datnq\}@qti.qualcomm.com}
}

\begin{document}
\maketitle
\begin{abstract}
Automatic Speech Recognition (ASR) performance is heavily dependent on the availability of large-scale, high-quality datasets. For low-resource languages, existing open-source ASR datasets often suffer from insufficient quality and inconsistent annotation, hindering the development of robust models. To address these challenges, we propose a novel and generalizable data aggregation and preprocessing pipeline designed to construct high-quality ASR datasets from diverse, potentially noisy, open-source sources. Our pipeline incorporates rigorous processing steps to ensure data diversity, balance, and the inclusion of crucial features like word-level timestamps. We demonstrate the effectiveness of our methodology by applying it to Vietnamese, resulting in a unified, high-quality 500-hour dataset that provides a foundation for training and evaluating state-of-the-art Vietnamese ASR systems. Our project page is available at \url{https://github.com/qualcomm-ai-research/PhoASR}.
\end{abstract}

\section{Introduction}

Recent advances in Automatic Speech Recognition (ASR) have demonstrated that pretraining on massive, weakly supervised or self-supervised corpora produces models with improved robustness and cross-domain generalization~\cite{wav2vec2,whisper}. However, when adapting these models to specific languages or downstream tasks, the benefits of scale alone begin to plateau. At this point, high-quality supervised datasets become essential for achieving further performance improvements~\cite{whisper}.

Prior efforts to build supervised ASR datasets have typically followed one of two approaches: scripted read speech~\cite{librispeech,wsj,vctk,vivos} or spontaneous speech collection~\cite{gigaspeech,the_people_speech,lsvsc}. Read speech datasets are typically recorded in controlled environments, which ensures high audio quality and accurate transcriptions. While this approach offers clear advantages in terms of data quality, it comes with  drawbacks: these datasets are expensive to produce and often lack the linguistic variability and prosodic richness characteristic of natural conversations. This limitation reduces their effectiveness for real-world ASR applications.

In contrast, datasets based on spontaneous speech offer several advantages: they are cheaper and more scalable to produce, and they better reflect real-world usage patterns due to their diverse sources, which often include publicly available content such as podcasts and YouTube videos. The creation pipeline for these datasets typically employs existing ASR models and forced alignment tools for automatic transcription, often supplemented by manual human review to ensure transcript quality~\cite{vimd,vietmed}.

For Vietnamese, existing open-source datasets exemplify these trade-offs. Some are based on controlled read speech but suffer from limited diversity~\cite{vivos}, while others draw from spontaneous sources but struggle with inconsistent transcription quality~\cite{vimd}. Furthermore, these datasets often vary in sampling rate and format, with preprocessing steps that are poorly documented~\cite{vsv1100}. This inconsistency makes it challenging to combine datasets or reuse them effectively in unified training setups.

Timestamps are rarely included in these datasets, which limits their use for training models that require fine-grained alignment—such as those used in subtitle generation or audio editing. While post hoc alignment tools like Montreal Forced Aligner~\cite{mfa} or other alignment models~\cite{wav2vec2} can be applied to ASR transcription outputs to provide timestamps~\cite{whisperx}, this approach increases computational cost and creates formatting challenges. These alignment tools often require transcripts to be in standard written form. For example, numbers must be spelled out (e.g., ``forty five'' instead of ``45'') to avoid out-of-vocabulary issues. However, ASR systems typically output digits (e.g., ``45'') rather than words (``forty five''), creating a mismatch with alignment model requirements. This incompatibility is problematic because the spelled-out format may not match what end-users actually want.

Similarly, post-processing models for punctuation restoration operate only on text without access to the original audio, preventing them from correctly placing punctuation based on acoustic cues like pauses or intonation.

To address these limitations, we propose a generalizable pipeline for aggregating and preprocessing ASR data from diverse sources. Our pipeline leverages existing deep learning models and auxiliary tools to clean, normalize, and align raw audio-text pairs, producing word-level timestamps and datasets that are high-quality, balanced, and feature-rich. We demonstrate this pipeline's effectiveness by applying it to Vietnamese, creating a 500-hour dataset suitable for fine-tuning and evaluating modern ASR systems. This pipeline can be adapted to different languages facing similar data quality challenges, making it an important contribution to advancing speech processing research.

\section{Related Work}

\subsection{Vietnamese ASR Datasets}

The landscape of Vietnamese ASR is shaped by a variety of datasets, each with its own strengths and limitations. These can be broadly categorized into read speech and spontaneous speech corpora. Read speech datasets include CMV-vi-14~\cite{cmv14}, VIVOS~\cite{vivos} and FPT Open Speech Dataset (FOSD)~\cite{fpt}. Spontaneous speech datasets include VSV-1100~\cite{vsv1100}, viVoice~\cite{vivoice},  BUD500~\cite{bud500}, VLSP 2020,\footnote{\url{https://vlsp.org.vn/vlsp2020/eval/asr}} Vietnam-Celeb~\cite{vietnamceleb}, ViMD~\cite{vimd}, LSVSC~\cite{lsvsc}, VietMed-L~\cite{vietmed}. In this section, we will briefly highlight some datasets from both categories.

\paragraph{Read speech datasets.} Common Voice~\cite{cmv14} is a large-scale, multilingual crowdsourced dataset initiative by Mozilla, aiming to provide free and publicly available voice data for speech technology development. Volunteers contribute by recording themselves reading sentences from a public-domain text corpus, and other users validate the recordings by listening to them. While it has grown to be one of the largest public voice datasets, covering over 100 languages, the data distribution is highly skewed. For many low-resource languages, the available data is very limited. For instance, the Vietnamese portion  {CMV-vi-14} of the dataset in the 14th version contains less than five hours of validated audio,\footnote{\url{https://huggingface.co/datasets/mozilla-foundation/common_voice_14_0}} which is insufficient for training high-performance ASR systems from scratch. The text in the dataset does include punctuation and casing, but timestamps are unavailable.

VIVOS~\cite{vivos} is a Vietnamese speech corpus originally created for text-to-speech research. It consists of 15 hours of read speech from a small number of speakers recorded in a controlled environment. While the audio quality is high, the provided transcripts often lack punctuation and capitalization and do not include word-level timestamps. These limitations make it less suitable for training models that require rich text features or precise audio-text alignment without further processing.

\begin{figure*}[t!]
    \centering
    \includegraphics[width=0.95\linewidth]{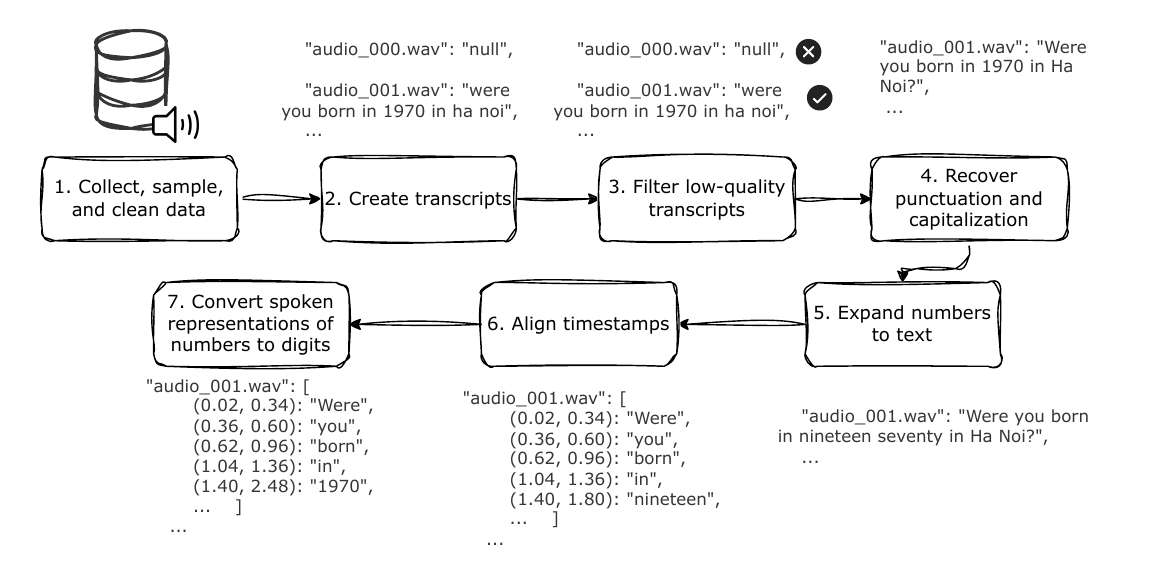}
    \caption{Overview of our pipeline for building a high-quality ASR dataset from multiple sources. The figure uses an English example for illustration. We apply this pipeline to Vietnamese. But, it is adaptable to other languages.
    }\label{fig:pipeline}
\end{figure*}

\paragraph{Spontaneous speech datasets.}
BUD500~\cite{bud500} is presented as a large corpus of spontaneous Vietnamese speech. However, limited information is publicly available regarding its data collection and annotation methodology, making it difficult to assess its quality comprehensively. Our analysis in Table~\ref{tab:dataset_sizes} shows that a substantial portion of the dataset fails to pass our quality filters, suggesting inaccuracies in the provided transcripts. Additionally, the dataset lacks standardized punctuation, capitalization, and word-level timestamps, which limits its direct applicability for training robust ASR models.
 
VietMed-L~\cite{vietmed} is a 16-hour labeled subset of the VietMed dataset for Vietnamese medical ASR. The dataset was created through a multi-stage, computer-assisted annotation workflow. The process began with initial transcripts generated by YouTube's automatic captioning service, which were then independently corrected by two native Vietnamese speakers. The two corrected versions were compared, and segments with significant transcription differences were excluded from the dataset. For final quality assurance, a small validation portion was manually transcribed from scratch by three annotators with medical backgrounds and subsequently merged with the computer-assisted versions. Although the dataset transcripts benefit from both human and machine annotations, they lack punctuation, capitalization, and timestamps.

\medskip

\noindent \textbf{Note that} other datasets in the read and spontaneous categories face similar challenges to those reviewed above: they lack punctuation, capitalization, and word-level timestamps.

\subsection{Forced Alignment for Timestamp Generation}

A key challenge in preparing ASR datasets is generating accurate word-level timestamps through a process known as \textbf{forced alignment}.

WhisperX \cite{whisperx} adopts a two-stage, model-based approach to forced alignment.
In the first stage, a large, pre-trained ASR model (by default, OpenAI's Whisper) generates a highly accurate transcript of the audio. In the second stage, the system performs forced alignment using a separate model, typically \texttt{wav2vec2} \cite{wav2vec2}, that has been fine-tuned specifically for alignment tasks. This alignment model employs a Connectionist Temporal Classification (CTC) head, which outputs a probability distribution over vocabulary tokens (e.g., characters or phonemes) for each time step of the audio representation. Given the transcript from the first stage, the aligner identifies the most probable sequence of token spikes in the CTC output matrix that corresponds to the transcript. The time steps associated with the start and end of each word's tokens are then merged to produce the final word-level timestamps.

This methodology bypasses the need for a pronunciation dictionary, making it highly adaptable to various languages and domains. However, the process is computationally intensive due to the use of large models, and the alignment step often requires transcripts to be in standard written form. For instance, numbers may need to be converted to their full textual representation (e.g., 45 to forty-five), creating an intermediate processing step that may not be desired in the final output. Our pipeline incorporates this approach while managing these complexities to achieve high-quality, feature-rich results.

\section{Our pipeline}\label{sec:method}

Our dataset creation pipeline, shown in Figure~\ref{fig:pipeline}, consists of the following steps:

\begin{enumerate}[leftmargin=*]
\setlength\itemsep{-0.1em}

    \item \textbf{Collect, sample, and clean data:}\ \  This step comprises: (i) collecting a diverse set of open-source audio samples from publicly available repositories, (ii) filtering out audio samples containing invalid characters or non-standard punctuation, and (iii) 
normalizing the audio volume and resampling each sample to a predefined target sampling rate to ensure consistency across the dataset.

    \item \textbf{Create transcripts:}\ \  For audio datasets lacking associated transcripts, this step comprises: (i) applying two state-of-the-art pre-trained ASR models to independently generate transcripts, (ii) computing the word error rate (WER) by comparing the transcript output of one ASR model against that of the other, and (iii) retaining only those audio-transcript pairs for which the computed WER falls below a predefined threshold, thus ensuring transcript reliability.

    \item \textbf{Filter low-quality transcripts:}\ \  For datasets that include pre-existing transcripts, this step involves: (i) applying an additional ASR model to generate a secondary transcript for each audio sample, (ii) comparing the secondary transcript with the provided transcript to compute the WER, (iii) discarding samples where the computed WER exceeds a predefined threshold, and (iv) using manually annotated transcripts without modification, where available.
    
    \item \textbf{Recover punctuation and capitalization:}\ \  To enhance transcript readability, this step includes: (i) training and deploying a neural network model to restore punctuation and capitalization in the transcripts, (ii) normalizing both the model input and output by converting them to lowercase and removing punctuation, and (iii) retaining only those samples for which the normalized input and output match exactly, thereby ensuring the fidelity of the restored formatting.

    \item \textbf{Expand numbers to text:}\ \  All numeric tokens in the transcripts are converted to their corresponding full-text (spoken) representations. For example, the numeral "30" is transformed into the word "thirty." This transformation is performed to ensure compatibility with a subsequent alignment model that requires textual input for accurate processing.

    \item \textbf{Align timestamps:}\ \  This step involves applying an alignment model to the audio signal and its corresponding transcript. The alignment model generates temporal metadata by assigning timestamps to each syllable in the transcript, thereby enabling word-level synchronization between the audio and the textual transcript.  
    
    \item \textbf{Convert spoken representations of numbers to digit:}\ \ This step is to convert the spoken textual representations of numbers back into their digit form. In instances where a single numeric expression spans multiple words and associated timestamps (e.g., ``one hundred twenty-three''), the corresponding timestamps are merged to represent the numeric entity as a unified temporal segment.

\end{enumerate}

This pipeline ensures that the resulting dataset is clean, standardized, and enriched with word-level timestamps, making it well-suited for training and evaluating modern ASR systems.

\begin{table*}[t!]
    \centering
    \resizebox{\textwidth}{!}{%
    \begin{tabular}{lrllllll}
        \toprule
        \textbf{Dataset} & \makecell[b]{\textbf{Sample}\\\textbf{Rate (Hz)}} & \makecell[b]{\textbf{Speech}\\\textbf{Type}} & \textbf{Domain} & \textbf{Text Length}  &  \textbf{Duration} &  \textbf{License}  \\
        \midrule
        VSV-1100 & 16000 & Spontaneous & General & 3--43 | 14 & 2--8 | 4 & Apache 2.0\\        viVoice & 24000 & Spontaneous & General & 1--60 | 16 & 1--19 | 4 & CC-BY-NC-SA 4.0\\
        BUD500 & 16000 & Spontaneous & General & 6--32 | 10 & 1--8 | 3 & CC-BY-NC-SA 4.0\\
        VLSP2020 & 16000 & Spontaneous & Unknown & 1--132 | 17 & 1--30 | 5 & Custom \\      
        Vietnam-Celeb & 16000 & Spontaneous & Interviews, Shows & W/o transcripts & 1--30 | 8 & Custom \\
        ViMD & 44100 & Spontaneous & Broadcast News & 6--132 | 62 & 2--30 | 19 & CC-BY-NC-ND 4.0\\
        LSVSC & 16000 & Spontaneous & General & 1--69 | 26 & 1--13 | 7 & CC-BY 4.0\\
        FOSD & 48000 & Read & Unknown & 1--59 | 12 & 3--6 | 4 & Custom \\
        VietMed-L & 8000 & Spontaneous & Medical & 5--39 | 24 & 2--10 | 7 & CC-BY-4.0 \\
        VIVOS & 16000 & Read & Unknown & 2--30 | 13 & 1--18 | 5 & CC-BY-NC-SA-4.0 \\
        CMV-vi-14 & 32000 & Read & General & 1--14 | 8 & 2--10 | 5 & CC0 \\
        \bottomrule
    \end{tabular}
    }
    \caption{Characteristics of the source datasets. "Text Length" and "Duration" are accounted for "words" and "seconds", respectively, with "min-max | mean" values. All datasets permit research use.}\label{tab:dataset_chars}    
\end{table*}

\section{Our PhoASR dataset}\label{sec:phoasr}

We demonstrate how the proposed pipeline is applied to Vietnamese to construct a 500-hour high-quality dataset PhoASR, as well as an extended version of its training set (PhoASR-3100h).

\subsection{PhoASR Dataset Creation}\label{ssec:refined_dataset}

\textbf{Step 1.}\ \ We download multiple datasets, including: VSV-1100~\cite{vsv1100}, viVoice~\cite{vivoice},  BUD500~\cite{bud500}, VLSP 2020\footnote{\url{https://vlsp.org.vn/vlsp2020/eval/asr}}, Vietnam-Celeb~\cite{vietnamceleb}, ViMD~\cite{vimd}, LSVSC~\cite{lsvsc}, FPT Open Speech Dataset (FOSD)~\cite{fpt}, VietMed-L~\cite{vietmed}, VIVOS~\cite{vivos}, CMV-vi-14~\cite{cmv14}, all of which are released under licenses that permit use for research purposes. Table~\ref{tab:dataset_chars} reveals considerable heterogeneity across the source datasets. Sampling rates vary widely from 8kHz to 48kHz, while the datasets exhibit diverse speech types ranging from read speech (CMV-vi-14, VIVOS) to spontaneous conversations (VSV-1100, VietMed-L) and mixed content. The domains are equally varied, covering general-purpose speech, medical conversations, broadcast news, and entertainment shows. However, some datasets lack proper documentation, with unknown domains and speech types.
Text lengths and audio durations also show significant variation, with ViMD containing the longest average samples (19s) and BUD500 the shortest (3s). While this heterogeneity presents significant preprocessing challenges, it results in a dataset whose diversity is crucial for enhancing the generalizability of any model trained on it. Table~\ref{tab:dataset_sizes} shows that the total duration of audio data is approximately 3342 hours, which includes 3171 hours of split training data.

To ensure dataset diversity and prevent dominance by particular speakers or dialects, we sample only portions of the downloaded datasets. For datasets with available speaker information, namely Vietnam-Celeb~\cite{vietnamceleb} and ViMD~\cite{vimd}, we limit each speaker to a maximum of 10 minutes. For the viVoice dataset~\cite{vivoice}, where audio is collected from YouTube, we treat the channel name as the speaker identifier and sample at most 30 minutes per channel. 
For VSV-1100~\cite{vsv1100} and BUD500~\cite{bud500}, which lack speaker information, we train a classification model to predict the province of each audio sample and limit sampling to 50 minutes per province. The classification model is trained using the ViMD~\cite{vimd} dataset, which includes province information. We use the \texttt{soxan} codebase~\footnote{\url{https://github.com/m3hrdadfi/soxan}} to finetune the \texttt{wav2vec2-base-vi}~\cite{wav2vec2_base_vi} model, achieving 43.48\% province prediction accuracy on the ViMD test set. Even with low classification accuracy, the model assigns different labels to different voices and groups similar voices together, ensuring speaker diversity in the sampling process. 
  
To ensure a consistent dataset, we perform cleaning and normalization. The data is first filtered by removing samples longer than 30 seconds and those with special characters in their transcripts, ensuring only standard punctuation (\texttt{,.!?}) and alphabet characters remain. All audio is then normalized in volume and resampled to 16 kHz using the \texttt{sox} library.\footnote{\url{https://sourceforge.net/projects/sox/}} This initial processing phase results in an 809-hour dataset.
  
\textbf{Step 2.}\ \ 
This step addresses datasets that lack pre-existing transcripts. Among the 11 datasets collected, Vietnam-Celeb~\cite{vietnamceleb} does not contain transcripts as it was constructed for speaker verification tasks. We use two state-of-the-art Vietnamese ASR models---\texttt{PhoWhisper-large}~\cite{phowhisper} and \texttt{ChunkFormer-large-vie}~\cite{chunkformer}---to predict the transcripts, then calculate the word error rate (WER) between their output predictions. Only audio samples with the corresponding WER lower than 5\% are retained.

\textbf{Step 3.}\ \ For datasets with pre-existing transcripts, we run \texttt{PhoWhisper-large} and compute the WER between the predicted transcript and the ground truth transcript provided by each dataset. Only samples with WER lower than 5\% are retained.

\begin{table*}[t!]
    \centering
    \begin{tabular}{l|rrr|rrr|r}
        \hline
        \textbf{Dataset} & \textbf{Original} & \textbf{Sampled} & \textbf{PhoASR} & \textbf{Train.} & \textbf{Valid.} & \textbf{Test} & \textbf{PhoASR-3100h} \\
        \hline
        VSV-1100 & 1144.53 & 48.85 & 25.69 & 25.69 & N/A & N/A & 1144.52\\
        viVoice & 1016.97 & 89.41 & 33.42 & 33.42 & N/A & N/A & 997.07\\
        BUD500 & 461.90 & 45.24 & 20.61 & 20.55 & 0.03 & 0.03 & 456.58\\
        VLSP2020 & 261.83 & 261.83 & 205.87 & 196.48 & 2.06 & 7.33 & 258.91\\
        Vietnam-Celeb & 187.37 & 98.95 & 37.37 & 37.37 & N/A & N/A  & 62.58\\
        ViMD & 102.56 & 97.57 & 61.60 & 53.13 & 6.54 & 1.93 & 98.10 \\
        LSVSC & 100.66 & 100.66 & 73.47 & 61.94 & 7.68 & 3.85 & 98.05\\
        FOSD & 30.18 & 30.18 & 21.97 & 20.15 & 1.08 & 0.74 & 30.12\\
        VietMed-L & 15.93 & 15.93 & 4.20 & 4.08 & N/A & 0.12 & 15.92 \\
        VIVOS & 15.66 & 15.66 & 14.31 & 12.88 & 0.91 & 0.52 & 15.66\\
        CMV-vi-14 & 4.79 & 4.79 & 4.15 & 2.91 & 0.39 & 0.85 & 4.79\\
        \hline
        \textbf{Total} & 3342.38 & 809.07 & 502.67 & 468.60 & 18.70 & 15.37 & 3100.52 \\
        \hline
    \end{tabular}%
    \caption{Dataset sizes (in hours) at different stages of the pipeline and the final split distribution. "Train." and "Valid." refer to the Training and Validation splits of PhoASR, respectively. \texttt{N/A} indicates that the validation or test split is not available for a given dataset. "PhoASR-3100h" denotes the extended training set, which combines minimally processed data with the high-quality data from the PhoASR training split (see details in Section~\ref{ssec:hybrid_dataset}).}\label{tab:dataset_sizes}
    
\end{table*}

\textbf{Step 4.}\ \  We fine-tune the \texttt{bartpho-syllable-base} model \cite{bartpho} for capitalization and punctuation recovery. The training data is taken from the "news-corpus",\footnote{\url{https://github.com/binhvq/news-corpus}} which contains main content texts from Vietnamese news articles. We select a subset of 5M samples without special characters and apply four transformation strategies to generate training variations: 85\% of samples are lower-cased and stripped of punctuation, 5\% have only punctuation removed, 5\% are only lower-cased, and the final 5\% are left unchanged. We then fine-tune the model for 20 epochs and use the fine-tuned one to predict punctuation and capitalization for the audio transcripts. To ensure prediction quality, we first lowercase both the fine-tuned model's transcript input and prediction output, then remove punctuation from both lowercased variants. We compare these processed versions of the input and output, retaining only samples where they are identical, ensuring the model has not added, removed, or altered any words during the capitalization and punctuation recovery process.

\textbf{Step 5.}\ \  Since our end goal is to build a dataset with timestamps, we need to run the transcripts through an alignment model. However, available alignment models require text to be in standard spoken form, which is not the case for our filtered transcripts. Therefore, we need to convert numbers in the transcripts to their text form. To accomplish this, we use a pre-trained model for text normalization \cite{phoaudiobook}.\footnote{\url{https://huggingface.co/thivux/PhoTextNormalization}} 
We compare the model's input and output to obtain the number-to-text mapping and replace the numbers in the transcripts with their word forms.

\textbf{Step 6.}\ \  With the transcripts in standard spoken form, we can now run the alignment model. We use \texttt{whisperx}~\cite{whisperx} as the alignment framework, coupled with a \texttt{wav2vec2}-based model fine-tuned on Vietnamese \cite{wav2vec2khanhld}, to generate word-level timestamps for the transcripts.  
To be compatible with the Whisper model, timestamp values obtained from \texttt{whisperx} are quantized into 20 ms (0.02 s) intervals---for example, <|0.00|>, <|0.02|>, ..., <|30.00|>---which serve as textual timestamp tokens during training.

\textbf{Step 7.}\ \ 
Converting the spoken textual representations of numbers back to the standard numerical form can be  done by taking the number-to-text mapping from Step 5 and switching the texts with their corresponding numbers. We then take the starting timestamp of the text's first word and the ending timestamp of the text's last word as the starting and ending timestamps of the numerical form.

\paragraph{Discussion.} The final result is a 502.67-hour high quality dataset (PhoASR).  
Table~\ref{tab:dataset_sizes} shows the size of the datasets at the start, after sampling, and after intensive filtering and refining through the pipeline. We can observe substantial data reduction through our pipeline, with about 502.67 / 809.07 $\simeq$ 62\% of the sampled data being retained after all processing steps. This substantial reduction highlights the prevalence of low-quality samples in the original datasets. Notably, medical domain data (VietMed-L) showed the highest reduction rate, retaining only 4.20 / 15.93 $\simeq$ 26\% of its sampled size, despite the authors' claim of a manual verification process. In contrast, read speech datasets like VIVOS and CMV-vi maintained over 90\% of their data, indicating their superior initial quality. This is expected, as in VIVOS and CMV-vi, the audio is recorded with people reading from prepared scripts, while in the rest of the datasets, the transcripts are generated by ASR models and thus are more prone to errors.

To investigate the balance of the dataset, we also add region information to each sample. Vietnamese has three main regional accents: Northern, Central, and Southern. We first train a dialect classifier using samples from raw datasets that contain region metadata: Vietnam-Celeb, ViMD, and LSVSC (389 hours in total). The \texttt{wav2vec2-base-vi}\cite{wav2vec2_base_vi} model is fine-tuned on this data for 10 epochs and achieves 90\% accuracy. This classifier is then used to predict the regional accent for all samples in our dataset. The results in Table \ref{tab:region_stats} show that Northern accent is the most dominant, followed by the Central accent, while the Southern accent is the least represented.

To assess the impact of our pipeline, we manually evaluated the quality of the original test set versus our refined PhoASR test set (containing 15.37 hours of audio). We sampled 20 examples from each component dataset, listened to the audio, created ground-truth transcripts, and compared them with the corresponding transcripts in the test sets. This manual check reveals that the average WER for the original test set is 2.73\%, with VietMed-L having the highest WER at 14.89\%, while the refined test has almost perfect transcripts, with an average WER of only \textbf{0.23}\%. 
Therefore, we use the refined test set for evaluation in our experiments. 

\begin{table}[t]
\centering

\setlength{\tabcolsep}{3pt}
\begin{tabularx}{\columnwidth}{Xllll} 
\toprule
\textbf{Region} & \textbf{Training} & \textbf{Validation} & \textbf{Test} & \textbf{Total} \\
\midrule
North    & 266.17  & 11.48  & 8.52  & 286.17 \\
South    & 138.09  & 4.01   & 4.37   & 146.47 \\
Central  & 64.34   & 3.21   & 2.48   & 70.03  \\
\midrule
\textbf{Total}    & 468.60  & 18.70  & 15.37  & 502.67 \\
\bottomrule
\end{tabularx}
\caption{Our dataset distribution by region and split. From this point onward, \textbf{`64h' and `469h' refer to 64.34 hours and 468.6 hours of audio}, respectively.}
\label{tab:region_stats}
\end{table}

\subsection{PhoASR-3100h Dataset Creation}\label{ssec:hybrid_dataset}

In addition to our high-quality PhoASR dataset, we construct a larger mixed dataset to investigate the benefits of combining rigorous processing with increased scale. This mixed dataset is created by applying a minimal processing---defined as initial cleaning (Step 1) followed by punctuation and capitalization recovery (Step 4)---to the 3171 hours of raw training data and then merging the result with our PhoASR training set of 469h data.

In particular, recall that a portion of the {Vietnam-Celeb} audios already have transcripts generated in Step 2 of our PhoASR-specific pipeline. For the remaining {Vietnam-Celeb} audios that lack transcripts, we use PhoWhisper-large to generate them. 
We then apply the aforementioned minimal processing to the entire 3171-hour training data. For any audio files that also appear in our PhoASR training set, we replace their original or generated transcripts with the corresponding high-quality timestamped versions from the PhoASR training set. This process yields the PhoASR-3100h training set, which combines the precision of our rigorous pipeline with the lexical diversity of larger-scale data.

Table~\ref{tab:dataset_sizes} presents the statistics of the PhoASR-3100h dataset. It represents nearly the entire training corpus (3100 out of 3171 hours), with a small amount of data excluded during the cleaning step.

\section{Experiments}\label{sec:experiments}

\subsection{Setup}

\noindent \textbf{Text Accuracy}: To measure the performance of text prediction, we use the Word Error Rate (WER) metric. WER is a standard metric for ASR that calculates the number of substitutions, deletions, and insertions between the predicted and reference texts, normalized by the total number of words in the reference text. A lower WER indicates better performance. We report two variants of WER~\cite{huggingface_wer}:

\begin{itemize}[leftmargin=10pt]
    \item \textbf{Orthographic WER (O-WER)}: This is calculated using the raw, unnormalized text, preserving the original capitalization and punctuation. It offers a strict measure of the model's ability to produce well-formatted output.
    \item \textbf{Normalized WER (N-WER)}: Before calculating WER, both the predicted and reference texts are normalized by converting them to lowercase and removing all punctuation. This approach emphasizes core lexical accuracy, disregarding formatting differences.
\end{itemize}

\noindent \textbf{Timestamp Accuracy}: To evaluate the accuracy of word-level timestamps, we use the following metrics:

\begin{itemize}[leftmargin=10pt]
    \item \textbf{F$_1$-score}: This is the primary metric for evaluating timing accuracy, derived from True Positives (TP), False Positives (FP), and False Negatives (FN). A predicted word is considered a TP if it matches a reference word in content and their temporal overlap falls within a predefined collar. An FP refers to a predicted word with no corresponding reference, while an FN is a reference word that the model fails to predict.
    \item \textbf{mean Intersection over Union (mIoU)}: This metric assesses localization accuracy. For each predicted word that matches a reference word, the Intersection over Union (IoU) of their timestamps is computed. If no match is found, a score of 0 is assigned. The final metric is the average IoU across all matched words, with higher values indicating better performance.
\end{itemize}

\noindent \textbf{Implementation:} Details of the implementation are provided in Appendix~\ref{sec:implementation_details}.

\subsection{Impact of Regional Accents}\label{ssec:accents}

We investigate the impact of regional accents on ASR performance. To ensure a balanced comparison, we sample 64h of training data--the total available from the Central region--from each of the Northern and Southern regions, resulting in a combined dataset of 64 $\times$ 3 = 192 hours (192h). We fine-tune the \texttt{whisper-small}~\cite{whisper} model for 40 epochs on the 192-hour dataset, as well as on each of the 64-hour regional subsets.

\begin{table}[t!]
    \centering
    \setlength{\tabcolsep}{4pt}
    \resizebox{1\columnwidth}{!}{%
    \begin{tabular}{l|l|cccc}
        \hline
        \multicolumn{2}{c|}{\textbf{Dataset}} & \textbf{North} & \textbf{Central} & \textbf{South} & \textbf{Overall} \\
        \hline
        \multirow{4}{*}{\rotatebox[origin=c]{90}{{O-WER}}} 
        & 64h-north & \underline{19.89} & 26.16 & 20.85 & \underline{21.08} \\
        & 64h-central & 21.66 & \underline{25.40} & 20.68 & 22.16 \\
        & 64h-south & 21.57 & 26.16 & \underline{19.17} & 21.72 \\
        & 192h & \textbf{15.88} & \textbf{19.86} & \textbf{15.81} & \textbf{16.53} \\
        \hline 
        \multirow{4}{*}{\rotatebox[origin=c]{90}{{N-WER}}} 
        & 64h-north & \underline{16.32} & 22.46 & 17.22 & \underline{17.54} \\
        & 64h-central & 17.62 & \underline{22.10} & 16.99 & 18.27 \\
        & 64h-south & 17.69 & 22.60 & \underline{15.77} & 18.02 \\
        & 192h & \textbf{12.18} & \textbf{16.56} & \textbf{12.36} & \textbf{12.97} \\
        \hline
    \end{tabular}
    }
    \caption{WER scores (\%) for regional subsets when fine-tuning \texttt{whisper-small} with 40 training epochs.}\label{tab:regional_wer}
\end{table}

We then evaluate each model on three regional subsets of the test set. As shown in Table \ref{tab:regional_wer}, the model trained on the combined 192-hour dataset achieves the best overall performance, highlighting the benefits of training on diverse accents. In contrast, models trained on a single accent perform well on their matched region but are less effective on others. Furthermore, the Central accent is consistently the most difficult for models trained on other accents, which may indicate greater linguistic and phonetic divergence of that dialect from the others.

\subsection{Timestamp Ratio}\label{ssec:timestamp_ratio}

In this experiment, we fine-tune \texttt{whisper-small} to generate both timestamp and transcript tokens. Our goal is to determine the optimal proportion of timestamped data in the training set. To this end, we fine-tune the model on the 100h subset for 40 epochs using different timestamp ratios: 0\%, 25\%, 50\%, 75\%, and 100\%. Here, the ratio refers to the proportion of timestamped data included in each training epoch.

Table~\ref{tab:timestamp_ratio} shows trade-offs between transcription and alignment performance across different timestamp ratios. While the 0\% ratio (no timestamps) achieves the best transcription accuracy (lowest O-WER and N-WER), the 100\% timestamp ratio delivers the highest timestamp alignment scores but at the cost of reduced transcription quality. The 50\% timestamp ratio provides the optimal balance, maintaining competitive transcription and timestamp alignment performances. 
This suggests a clear trade-off in the multi-task learning setup: incorporating timestamp prediction improves alignment capabilities but can impair transcription accuracy when overemphasized.

\begin{table}
    \centering
    \setlength{\tabcolsep}{4pt}
    \resizebox{1\columnwidth}{!}{%
    \begin{tabular}{lcccc}
        \toprule
        \textbf{Ratio} & \textbf{O-WER} $\downarrow$ & \textbf{N-WER} $\downarrow$ & \textbf{F1} $\uparrow$ & \textbf{IoU} $\uparrow$ \\
        \midrule
        0\% & \textbf{19.21} & \textbf{15.79} & 75.83 & 49.26 \\

        25\% & 22.66 & 19.23 & 76.42 & 49.65 \\

        50\% & \underline{20.40} & \underline{16.84} & 76.72 & 51.25 \\

        75\% & 22.16 & 18.38 & \underline{79.45} & \underline{54.84} \\ 

        100\% & 21.88 & 17.59 & \textbf{80.61} & \textbf{57.21} \\
        \bottomrule
    \end{tabular}%
    }
    \caption{Performance results when fine-tuning \texttt{whisper-small} with different timestamp ratios (\%) on 100h training subset for 40 training epochs.}\label{tab:timestamp_ratio}
\end{table}

\subsection{Different Models}\label{ssect:models}

We compare Whisper with another leading ASR model, \texttt{wav2vec2}~\cite{wav2vec2}, by fine-tuning \texttt{whisper-small} and \texttt{wav2vec2-xls-r-300m} on our 469h training set for 40 epochs. In this setup, \texttt{whisper-small} is trained to generate both timestamp and transcript tokens using a 50\% timestamp ratio, whereas \texttt{wav2vec2-xls-r-300m} is trained to generate transcript tokens only. This results in two fine-tuned models, \texttt{PhoASR-whisper-small-469h} and \texttt{wav2vec2 (469h)}, respectively. 

Table~\ref{tab:model_comparison} shows that \texttt{PhoASR-whisper-small-469h} consistently outperforms \texttt{wav2vec2} across all evaluation metrics. In particular, \texttt{PhoASR-whisper-small-469h} achieves an O-WER of 12.46\% and an N-WER of 8.69\%, which are substantially better than those of \texttt{wav2vec2} (51.15\% O-WER and 14.10\% N-WER). Furthermore, fine-tuning yields a massive improvement over the pre-trained \texttt{whisper-small} baseline, reducing O-WER from 70.16\% to 12.46\% and N-WER from 64.07\% to 8.69\%, confirming the necessity of domain adaptation for Vietnamese.

A comparison between \texttt{PhoWhisper-small}, which was fine-tuned from \texttt{whisper-small} on a larger 844h dataset, and our model \texttt{PhoASR-whisper-small-469h} highlights the superiority of our high-quality dataset. Our model achieves a better N-WER (8.69\% vs. 8.97\%) and a notably lower O-WER (12.46\% vs. 33.90\%). This result underscores the importance of data quality over quantity; despite being trained on nearly half the data, our model produces more accurate and better-formatted transcripts, demonstrating the effectiveness of our data processing pipeline.

\begin{table}[t!]
    \centering
    \setlength{\tabcolsep}{3pt}
    \resizebox{1\columnwidth}{!}{%
    \begin{tabular}{lcccc}
        \toprule
        \textbf{Model} & \textbf{O-WER} $\downarrow$ & \textbf{N-WER} $\downarrow$ & \textbf{F1} $\uparrow$ & \textbf{IoU} $\uparrow$ \\
        \midrule
        whisper-small (Pre-trained) & 70.16 & 64.07 & - & -\\
        ChunkFormer-large-vi (25K) & 32.43 & \textbf{6.89} & - & -\\
        PhoWhisper-small (844h) & 33.90 & 8.97 & - & -\\
        wav2vec2 (469h) & 51.15 & 14.10 & - & -\\
        PhoASR-whisper-small-469h & \underline{12.46} & 8.69 & \underline{76.40} & \underline{55.62} \\
        \textbf{PhoASR-whisper-small-3100h} & \textbf{11.70} & \underline{8.20} & \textbf{83.68}  & \textbf{57.39} \\
        \bottomrule
    \end{tabular}
    }
    \caption{Obtained scores for different models: "PhoASR-whisper-small-469h" trained for 40 epochs; "PhoASR-whisper-small-3100h" trained for 15 epochs. See Appendix~\ref{sec:appendix_results} for results on each component dataset. ChunkFormer was trained on an internal non-public dataset of \textit{25K} audio hours for \textit{200} epochs. PhoWhisper was trained on a dataset of \textit{844} audio hours, in which \textit{586} hours are private data.}\label{tab:model_comparison}
\end{table}

\subsection{Scaling with PhoASR-3100h}\label{ssec:scaling}

While Section \ref{ssect:models} demonstrates that rigorous processing allows smaller datasets to outperform larger noisy ones, massive scale remains crucial for covering long-tail vocabulary and diverse acoustic conditions. To combine the precision of our high-quality data with the diversity of large-scale pre-training, we evaluate the scaling potential by training \texttt{whisper-small} on our PhoASR-3100h mixed dataset (described in Section \ref{ssec:hybrid_dataset}) for 15 epochs, resulting in \texttt{PhoASR-whisper-small-3100h}.

Table~\ref{tab:model_comparison} presents the evaluation results. As expected, adding more training data gives a boost to the performance. \texttt{PhoASR-whisper-small-3100h} improves across all metrics compared to \texttt{PhoASR-whisper-small-469h}. For example, it reduces the O-WER from 12.46\% to 11.70\% and improves timestamp prediction score F1 from 76.40\% to 83.68\%.

Compared to other baseline models, \texttt{PhoASR-whisper-small-3100h} demonstrates a strong balance between lexical accuracy and orthographic correctness. For instance, \texttt{ChunkFormer-large-vi} achieves a lower N-WER of 6.89\%, benefiting from its massive 25,000-hour training set. However, its O-WER of 32.43\% is notably higher than \texttt{PhoASR-whisper-small-3100h}'s 11.70\%, indicating poor performance on punctuation and capitalization. In practical applications where transcripts are expected to be immediately usable, such as in automatic subtitling or meeting transcription, O-WER is a more critical metric than N-WER. A lower O-WER signifies that the output is well-formatted and requires minimal to no post-editing, making \texttt{PhoASR-whisper-small-3100h} a more suitable choice for such use cases.

\section{Pipeline Component Analysis}

We analyze the contribution of specific pipeline stages to the final model performance:

\noindent \textbf{Filtering \& Cleaning (Steps 1-3):} Our rigorous filtering allows a smaller, cleaner dataset to outperform a larger, noisier one. As shown in Table \ref{tab:model_comparison}, \texttt{PhoASR-whisper-small-469h} achieves a lower N-WER (8.69\%) compared to \texttt{PhoWhisper-small} (8.97\%), despite being trained on nearly half the amount of data (469h vs. 844h). This confirms that quality filtering is more effective than raw data quantity for lexical accuracy.

\noindent \textbf{Punctuation \& Capitalization (Step 4):} The impact of this step is evident in the O-WER metric. \texttt{PhoWhisper-small} exhibits a high O-WER of 33.90\%, while \texttt{PhoASR-whisper-small-469h} achieves 12.46\%. This significant gap highlights the necessity of explicit restoration (Step 4) for producing ready-to-use transcripts with correct formatting.

\noindent \textbf{Timestamp Alignment (Steps 5-7):} The results in Section \ref{ssec:timestamp_ratio} (Table \ref{tab:timestamp_ratio}) serve as an ablation for the alignment steps. Increasing the timestamp ratio from 0\% to 100\% directly correlates with improved alignment performance (F1 and mIoU), confirming that our forced alignment process effectively imparts fine-grained temporal understanding to the model.

\section{Conclusion}\label{sec:conclusion}

We present a pipeline for creating high-quality ASR datasets from noisy, open-source audio. Our method produces clean transcripts with reliable timestamps, punctuation, and capitalization, which eliminates the need for separate post-processing tools. To demonstrate its effectiveness, we construct a 500-hour Vietnamese corpus and demonstrate its potential as a strong foundation for both fine-tuning and benchmarking ASR models. While this pipeline was applied to Vietnamese, it is adaptable and practical for enhancing speech recognition across a broader range of languages.

\section*{Limitations}
While our pipeline substantially improves data quality, the filtering process itself might introduce bias. By selectively retaining samples that are cleanly processed by our toolchain, we may unintentionally favor certain acoustic environments, speaking styles, or accents that upstream models handle more effectively. As a result, the final dataset—though high in quality—may be less representative of the full diversity of real-world speech. Our experiments with the PhoASR-3100h dataset suggest that combining high-quality filtered data with larger, minimally processed corpora may be the most effective strategy for training robust and generalizable ASR models.

\bibliography{references}

\newpage

\appendix

\begin{table*}[t]
    \centering
    \resizebox{\textwidth}{!}{%
    \begin{tabular}{llrrrrrrrrr}
        \toprule
        \textbf{Metric} & \textbf{Model (Training Hours)} & \textbf{BUD500} & \textbf{VLSP2020} & \textbf{ViMD} & \textbf{LSVSC} & \textbf{FOSD} & \textbf{VietMed-L} & \textbf{VIVOS} & \textbf{CMV-vi-14} & \textbf{Overall} \\
                \midrule
        \multirow{4}{*}{\textbf{O-WER}}
        & ChunkFormer-large-vi (25K) & 18.22 & 38.25 & 13.13 & 14.16 & 18.9 & \underline{14.63} & 15.38 & 22.44 & 32.43 \\
        & PhoWhisper-small (844h) & 19.98 & 39.77 & 13.95 & 15.32 & 21.0 & 16.58 & 16.54 & 23.87 & 33.90 \\
        & wav2vec2 (469h)        & 21.76  & 60.89    & 17.44 & 19.45 & 28.79 & 22.23     & 33.45 & 31.04     & 51.15 \\
        & PhoASR-whisper-small-469h   & \underline{8.27} & \underline{14.25} & \underline{8.75} & \underline{6.72} & \underline{7.73} & \textbf{14.41} & \underline{5.92} & \underline{9.18} & \underline{12.46} \\
        & PhoASR-whisper-small-3100h  & \textbf{4.22} & \textbf{13.64} & \textbf{8.08} & \textbf{6.4} & \textbf{5.53} & 15.46 & \textbf{3.06} & \textbf{7.02} & \textbf{11.7} \\
        \midrule
        \multirow{4}{*}{\textbf{N-WER}}
        & ChunkFormer-large-vi (25K) & 3.03 & \textbf{8.71} & 4.46 & \textbf{1.19} & \textbf{0.79} & \textbf{5.95} & \textbf{1.37} & \textbf{2.97} & \textbf{6.89} \\
        & PhoWhisper-small (844h) & 6.59 & 10.95 & 5.55 & 2.64 & 3.36 & 7.9 & 2.69 & 4.81 & 8.97 \\
        & wav2vec2 (469h)        & \underline{2.42}   & 17.51    & 7.34  & 2.84  & 4.51  & 13.93     & 4.61  & 7.51      & 14.10 \\
        & PhoASR-whisper-small-469h   & 3.97 & 10.38 & \underline{4.64} & 2.64 & 4.58 & \underline{9.51} & 4.33 & 6.34 & 8.69 \\
        & PhoASR-whisper-small-3100h  & \textbf{0.92} & \underline{10.09} & \textbf{3.99} & \underline{2.49} & \underline{2.35} & 10.2 & \underline{1.92} & \underline{4.35} & \underline{8.2} \\

        \midrule
        \multirow{2}{*}{\textbf{F1}}
        & PhoASR-whisper-small-469h   & \underline{77.31} & \underline{73.55} & \underline{73.77} & \underline{83.10} & \underline{75.69} & \underline{74.08} & \underline{76.25} & \underline{67.09} & \underline{76.40} \\
        & PhoASR-whisper-small-3100h  & \textbf{85.71} & \textbf{82.27} & \textbf{79.46} & \textbf{89.65} & \textbf{81.05} & \textbf{77.60} & \textbf{78.70} & \textbf{72.36} & \textbf{83.68} \\
        \midrule
        \multirow{2}{*}{\textbf{mIoU}}
        & PhoASR-whisper-small-469h   & \textbf{48.90} & \underline{58.06} & \underline{45.51} & \underline{51.32} & \underline{49.63} & \underline{47.90} & \underline{48.47} & \underline{46.30} & \underline{55.62} \\
        & PhoASR-whisper-small-3100h  & \underline{48.03} & \textbf{58.64} & \textbf{50.85} & \textbf{57.05} & \textbf{54.54} & \textbf{48.85} & \textbf{52.47} & \textbf{48.69} & \textbf{57.39} \\
        \bottomrule
    \end{tabular}%
    }
    \caption{O-WER (\%), N-WER (\%), F1 (\%), IoU (\%) for different models on each dataset within our refined  test set.}
    \label{tab:model_comparison_detailed_filtered}
\end{table*}

\section{Implementation details}\label{sec:implementation_details}

For our experiments, we fine-tuned two publicly available pre-trained models: \texttt{whisper-small}\footnote{\url{https://huggingface.co/openai/whisper-small}} and \texttt{wav2vec2-xls-r-300m}.\footnote{\url{https://huggingface.co/facebook/wav2vec2-xls-r-300m}}

The fine-tuning settings for \texttt{whisper-small} are  kept consistent across all experiments. We use a peak learning rate of $1.25 \times 10^{-5}$ with an AdamW optimizer and a linear learning rate scheduler. For \texttt{wav2vec2-xls-r-300m}, we use a peak learning rate of $1 \times 10^{-4}$ and train it with the Connectionist Temporal Classification (CTC) loss function. To improve generalization, we apply SpecAugment with a time masking probability of 0.75, a feature masking probability of 0.25, and a feature mask length of 64.

All models are trained for 40 epochs, with the first 5,000 steps used for warm-up, except for \texttt{whisper-small} trained on the PhoASR-3100h dataset for 15 epochs due to resource constraints. All experiments are conducted on a system with 4 NVIDIA A100  40GB GPUs. We use a per-device batch size of 4 and 4 gradient accumulation steps, which result in an effective global batch size of 64. The best-performing checkpoint for each model is selected based on the lowest N-WER achieved on the PhoASR validation set. We then employ the selected checkpoint to report final performance results on the PhoASR test set.

\section{Detailed Evaluation Results}\label{sec:appendix_results}

This section contains the detailed evaluation results of different models on each dataset of our test set.

Table~\ref{tab:model_comparison_detailed_filtered} shows the performance on our test set. \texttt{PhoASR-whisper-small-3100h} consistently achieves the best O-WER, F1 and mIoU, demonstrating the benefits of a larger training corpus. Notably, the \texttt{VLSP2020} dataset proves to be the most challenging for all models. Furthermore, while \texttt{wav2vec2} achieves competitive N-WER scores, its O-WER is considerably large, suggesting a weaker performance in predicting correct capitalization and punctuation compared to the \texttt{whisper}-based models.

\end{document}